# A Synthesis of Logical and Probabilistic Reasoning for Program Understanding and Debugging

Lisa J. Burnell
Computer Science Engineering
The University of Texas at Arlington
and
American Airlines Knowledge Systems
DFW Airport, TX 75261-9616

Eric J. Horvitz*
Palo Alto Laboratory
Rockwell International Science Center
444 High Street
Palo Alto, CA 94301

## Abstract

We describe the integration of logical and uncertain reasoning methods to identify the likely source and location of software problems. To date, software engineers have had few tools for identifying the sources of error in complex software packages. We describe a method for diagnosing software problems through combining logical and uncertain-reasoning analyses. Our preliminary results suggest that such methods can be of value in directing the attention of software engineers to paths of an algorithm that have the highest likelihood of harboring a programming error.

## 1 INTRODUCTION

We describe the integration of logical and uncertain reasoning methods to reason about the likely source and location of software problems. In particular, the methods have application to problems with maintaining and refining a large, complex piece of software that is used and refined over many years. Such *corporate legacy* software typically has a long history of evolution, growing and changing with contributions from many software engineers over time. Problems may be detected years after a software update or modification. In many cases, the software is poorly documented and incompletely tested. Frequently, people charged with the task of debugging find that engineers responsible for particular portions of program code have long since been promoted to other positions, or have left the company.

We have experimented with the application of uncertain reasoning to software debugging as part of the DAACS project, a doctoral research effort at the University of Texas at Arlington, and an ongoing project at the SABRE Knowledge Systems Group of American Airlines. DAACS (for Dump Analysis and Consulting System) is a program developed to assist software engineers to determine sources of software errors in SABRE, the most widely used time-shared reservation system in the world. Complex interactions by clients with the SABRE system, and continuing software development without complete verification, lead to intermittent problems that are recorded internally for later evaluation. In earlier work (Burnell and Talbot, 1993), we explored the use of a belief network for computing the likelihoods of alternative high-level explanations for software errors, based on a global analysis of the values of important variables in a program trace. In this paper, we describe a new approach being taken in DAACS-II that integrates the output of a logical analysis of feasible paths with probabilistic analyses of each path so as to prioritize the efforts of a software engineer in exploring different program paths to debug SABRE system software.

## 2 PROBLEM FOCUS

An autonomous program debugger must analyze a program to gain information about its behavior and identify the cause or causes of some set of faults. We have concentrated on applying logical and uncertain reasoning methods for interpreting the source of an abnormal termination of mainframe assembler language programs. *Abnormal termination* occurs when a program or subprogram in a timeshared environment violates an operating-system constraint. For example, an attempt may have been made by a process to reference an area of memory that is not allocated to that process. When an operating system detects such a violation, data are collected and formatted in a trace of recent history called a *memory dumpfile*, or *dump*. Engineers typically debug complex mainframe assembler language problems by poring over dumpfiles from programs that terminate abnormally.

A typical dumpfile contains a snapshot of relevant parts of memory at the time that an illegal operation occurred. The trace includes information about register contents, the processor status word, the program counter, the program object code, the memory location, and contents of program data. Software engineers, armed with a memory dump and a program

*Currently at Decision Theory Group, Microsoft Research Labs.

listing, pursue the source of a problem by identifying the specific instruction that led directly to the program termination, and then trace backwards through the operation of the program along execution paths in search of the principle cause of the termination. A valid *execution path* is a sequence of program instructions that *could have been* executed, depending on the outcome of *conditional branch* instructions. Along the way, several other key pieces of data from the memory dump may be examined to provide clues.

Debugging is intrinsically a problem of reasoning under uncertainty because the snapshot provided by information from a memory dump, and a recent history of the program operation before the system encountered an illegal operation, is an *incomplete* description of the software. In particular, we are uncertain about the identity of the execution path that was taken to reach the error. Typically many alternative paths are candidates and, thus, much of the effort of debugging is expended on determining the failing path.

As highlighted in Figure 1, our work on decision-support for debugging has two components:

1. Logical analysis to identify possible execution paths, and
2. Probabilistic analysis using a belief network for software failure to sort paths by the likelihood that the error resides on the paths, as well as to identify the likelihoods of alternative problems on each path.

The output of the analysis is an ordering over paths by likelihood, and a subdivision of the task of analyzing each path into an ordering over the likelihood of alternative classes of problem. Such orderings can speed the debugging process by minimizing the number of paths that must be explored before a problem is identified.

## 3   OPERATING SYSTEMS AND PROGRAM UNDERSTANDING

Our work on identifying execution paths draws upon techniques developed in work on *program understanding*. Program understanding tasks vary depending on the complexity of the processes being analyzed, and the level of abstraction available for debugging. For example, large, time-shared assembler programs provide challenges that are not usually encountered in microcomputer-based programs that are compiled from higher-level structured languages. With large assembler programs, variables are pointers that can access any part of memory and the code is unstructured. Also, input—output, execution traces, and intended behavioral descriptions are typically unavailable and are difficult to derive. Therefore, engineers must examine program structure and infer intended program behavior.

There has been ongoing related work on automated program understanding (APU). APU programs build abstract representations of a program and related information to facilitate reasoning (see, for example, (Arbon et al., 1992; Hartman, 1992; Kozaczynski et al., 1991; Lenz and Saelens, 1991; Selfridge, 1991). Most program understanding systems seek to match portions of programs to prototypical implementation plans. Applications pursued by APU researchers include student programming tutors, design recovery and reuse of software, and program language translation. In most APU projects, designers have sought to develop methods to interpret and understand an entire program segment. In the realm of automated program debuggers, the APU task involves understanding just enough about program behavior to determine failures in the code. In theory, this task is not as ambitious as comprehensive program understanding. However, unlike some of the work on program understanding, this and several other debugging projects concentrate on solving real-world problems with complex software, far from the realm of toy problems that are typically explored in research settings. Our system must operate on the software as it is written and can draw only on the information available in the existing memory dump. As mentioned earlier, the software that we seek to understand is unstructured assembler code that has been modified over several decades and that has never been tested comprehensively.

The DAACS system begins an analysis of a memory dump by searching for erroneous execution paths in assembler programs running under the IBM TPF[1] operating system. In response to an input request (*e.g.*, for a list of airline fares), a collection of program segments are activated to process the request. For this paper, we shall focus on problems with program segments attempting to reference pages of memory that are protected by other processes.

As background, each segment may be thought of as a function or subroutine within a traditional program. At activation, all of the segments are loaded in memory and a structure called the *entry control block* (ECB) is created in memory. All segments can access this memory. The ECB contains control and status information, and a small amount of storage for a work area (scratch data). Sixteen fixed-size areas of memory, called *core blocks*, may be dynamically requested or freed by any of the program segments. The location, size and status (*free* versus *held*) of all core blocks is recorded in the ECB. Working memory is divided into alternating protected and unprotected pages, indicated by a *protect key* of 0 and 1, respectively. All application segments use pages that are labelled as available to processes by having protect keys set to 1. As working memory is filled, the operating system makes new protected areas available by setting the segment protect keys to 1 when the memory is free for use. Free pages of mem-ory may be reprotected by other processes. An illegal memory reference occurs when a segment attempts to reference a protected page.

---

[1]TPF is a widely used operating system designed for transaction processing.

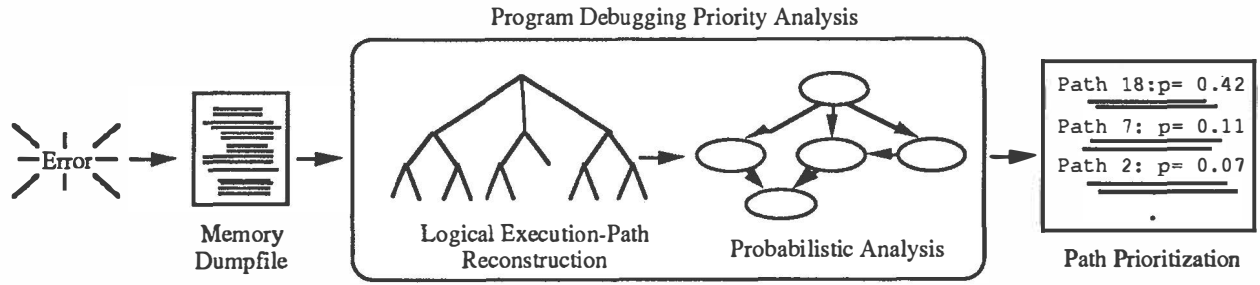

Figure 1: Overview of logical and uncertain reasoning approach. A memory dumpfile generated at the time of an error is analyzed to generate a graph of feasible execution paths. Each path is analyzed with a belief network to generate a list of paths prioritized by the likelihood that a path harbors an error.

## 4  LOGICAL ANALYSIS OF EXECUTION PATHS

Our goal is to determine the *root error* in a program segment. The root error is the initial instruction, or sequence of instructions upstream in a segment, that led to an erroneous situation that was *detected* by the operating system. For example, a common, detectable error for many programming languages and operating systems occurs when an instruction attempts to manipulate memory that was previously allocated to another process. For example, a variable in one process might point to a location that is outside a legal range of memory. This type of presentation is called a PASTHELD error.

### 4.1  Methods

To generate a set of feasible execution paths, we must backtrack from an error. We can represent a program as a directed graph where nodes are basic blocks of code and arcs represent the flow of execution control from a parent block to child blocks. Such graphs allow us to trace back along execution paths. In operation, only a single execution path is taken by the computer. Unfortunately, we usually cannot identify the single path taken before an error occurred. The values of variables in program segments can be assigned by actions defined in program segments that are outside the scope of the local segment analysis. We must consider loops and *transfer points*—commands that *branch* the possible flow of control from one basic block to several others based on the values of variables that are dynamically set by processes which lay outside the scope of an analysis. Thus, we must analyze blocks, looping, and branching instructions to identify and investigate feasible execution paths.

During the path-identification phase, we can attempt to refute or rule-out paths, via a logical analysis of components of the structure and triggers in a program. Many execution paths in a program may be pruned because they can be proven impossible based on the program structure and information that is available about the data input to a process.

Further simplification is possible because we need only to examine *partial paths*. For certain classes of errors, the first step in determining why an error occurred at an instruction is to examine the operands and their values to determine which variable (operand or constituent register) is clearly out-of-bounds. A trace backwards along a feasible execuation path from the point of the error terminates when an out-of-bounds variable is discovered along the path between the error and the first operation in a block that designates the value of the variable. In such cases, we know that the problem occurred either at the instruction where the variable was assigned or somewhere along the path between the assignment instruction and the error.

The logical reasoning component of DAACS employs such deterministic reasoning to generate a directed graph representing possible execution paths that flow to a detected memory-pointer error. The system then applies logical analysis to do path pruning. We have found that pruning is computationally feasible because the number of paths is generally small (less than 200 paths) and the task can be done quickly.

### 4.2  An Example

Let us consider a real debugging situation. We are alerted that a problem has been detected by the operating system running a critical reservations system: A pointer variable, VAR1, had been pointing to an area of memory that was allocated to another process. We note that this is a PASTHELD error, as VAR is pointing to a location that is 80 bytes outside a legal strip of memory. Assume that we are given a memory trace that contains the program code, the instruction at which an illegal memory reference occurred, and the values of relevant variables at the time the error occurred. Our goal is to determine the root error, the initial instruction upstream in the program that led to the error that was detected by the operating system.

```
Program Start
              °
              °
BLOCK A:

 1.  set var3=var4 (set outside segment)
 2.  set var1=var2 (set outside segment)
 3.  if var3 = 0 goto label y

BLOCK B:

 4.  label x:
              °
              °
 5.  set var1 = var1 + 80
              °
              °
 6.  set var3 = var3 - 1
 7.  if var3 = 0 goto label x

BLOCK C:

 8.  label y: set var2 = some value
 9.  set var3 = some value
10.  set var4 = some value
11.  print  var1<<  ***  ERROR  ***
              °
              °
```

Figure 2: Execution path identification example showing a block of instructions that are executed in a local sequence.

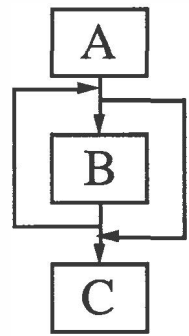

Figure 3: A control-flow graph of the instructions displayed in Figure 2.

In this case, the root error is the out-of-bounds value assigned to VAR1.

Given the location of the error and knowledge about the program, we begin to examine the structure of the execution paths. A partial listing of a program is shown in Figure 2. Figure 3 displays a control-flow graph of these instructions. Three basic blocks of code are displayed. A *basic block* is a set of instructions that is executed in a sequence. The only entrance to a basic block is to the first instruction, and the only exit from a basic block (to other blocks) is via the last instruction in that block.

The DAACS logical path-analysis component autonomously identifies multiple execution paths from the program statements given in Figure 2. Let us represent a component of an execution path from a parent block to a child block as Parent–Child. In doing a program tracing, DAACS notices the branching statement (a *goto* command) at the end of BLOCK A. Because it cannot determine the value that the branching statement was acting on in this case, it must consider two paths. In one execution path (path 1), DAACS notes the path BLOCK A—BLOCK B—BLOCK C with Block B being executed one or more times. DAACS also notes another execution path (path 2) is: BLOCK A—BLOCK C. With this VAR2 is reset, so we do not know what its value was at instruction 2. Moreover, VAR3 and VAR4 are reset, so their values are unknown too. A certain diagnosis will not be possible. DAACS can employ such control flow analysis to build a graph of execution paths that show possible channels for errors through a program. Once we have identified paths, we can attempt to refute paths with a logical analysis.

## 5  UNCERTAINTY ABOUT ERRORS ON PATHS

We have explored the application of probabilistic inference to autonomously evaluate and order executions paths for further study by software engineers. In this work, we apply expert knowledge about the probabilistic relationships between structural and numerical information within each path, and work to assign a probability that the problem resides within each path. For each path, we also report the relative likelihoods of alternative classes of problem. We have made several assumptions at this phase of our research. First, we make the typically valid assumption that problems leading to observations about invalid memory access are located on a single path. Second, we assume that all paths are equally likely to be the source of the error. We shall discuss work on the relaxation of this assumption in Section 6. Third, we assume that the paths identified by the logical analysis serve as the complete set of possibilities. Although this is typically valid, in some cases, we may not generate all valid paths. In such cases, the priority to exploring the paths assigned by the likelihoods will still be correct, given our other assumptions.

We have constructed distinct belief networks for key types of problems that can be detected by an operating system. The belief networks are constructed for application to each path identified by the logical analysis to provide an ordering over the paths for prioritizing the efforts of a software engineer. For each path on a list ordered by likelihood, the networks provide a subdivision assignment of the likelihoods of alternative classes of problem, should the problem be on that path.

The belief networks represent uncertain relationships about the nature and structure of a path and the likelihood that an error resides on the path, given that the path was taken by program. As depicted in Figure 4(b), each application of a belief network to a path, relevant to the key problem detected by the operating system, allows us to do Bayesian reasoning about the probability of alternate forms of error on that path.

We assess probabilities from expert engineers (or acquire data from debugging experience) of the form: $p(path\ findings\ |\ path\ error\ class)$. The system computes probabilities $p(path\ error\ class|path\ findings)$. For $n$ paths identified, we assume prior probabilities $p(path|n) = \frac{1}{n}$. Our uniform prior assumption is valid, given our reliance solely on the deterministic path analysis provided by the logical reasoner.

In practice, the logical component of DAACS generates all of the execution paths that could have led to an error detected by the operating system. In the uncertainty component of the analysis, prior probabilities are generated as a function of the number of paths, and each path is analyzed by the belief network relevant to the main error. After all paths have been analyzed and probabilities are renormalized, an ordering over the paths is output to engineers to direct their attention to paths in order of likelihood. For each path, information is relayed to engineers about the likelihood of different problems on the path, conditioned on the path being the source of the problem.

Consider the example of a network built to examine PASTHELD problems. The PASTHELD network computes, for each path, likelihoods of the form $p(error\ class\ on\ path|path\text{-}structure\ findings)$ for classes of error that describe different ways that an inappropriate value was assigned to an offending variable:

- BAD SET: Instructions in the segment have improperly initialized the erroneous variable.

- BAD ADJUST: Instructions in the segment have improperly modified a value stored in the variable (e.g. addition, multiplication, shift instructions)

- ENTERED BAD: The variable is not modified on the path, and so was assigned an invalid value before entering the segment being analyzed.

- BAD LOOP: There are errors with loop index initializations, adjustments, and exit tests that led to the erroneous value of the variable.

The PASTHELD belief network considers several key pieces of evidence that are associated with the existence of alternative faults on a path. Evidence collected from a path include information about:

- SYNTACTIC STRUCTURE: The structure of a path (e.g., structure SET FAIL rules out BAD ADJUST errors because no adjustment instructions exist on the path).

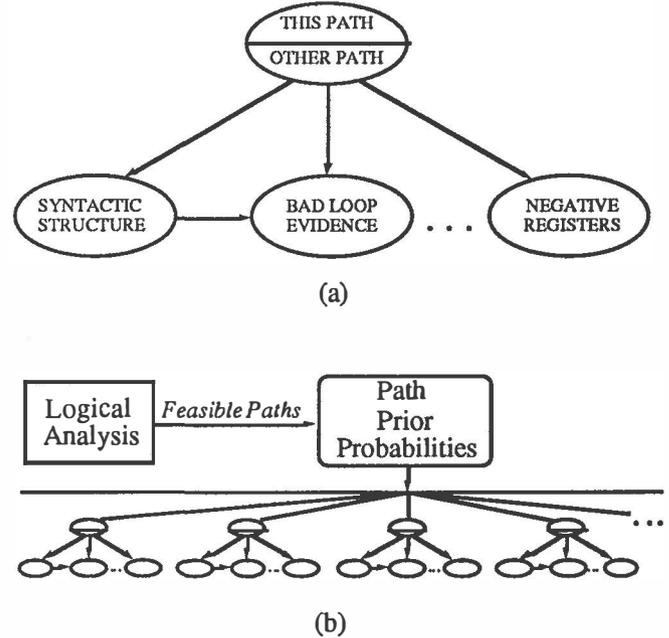

Figure 4: Set of belief networks used to evaluate paths. (a) The belief network for analyzing the probability that a problem lies on a path. (b) Analysis of multiple feasible paths of a program that have been identified by a deterministic analysis of alternative routes to an error.

- CLOSE REGS: The proximity of values of variables to the value assigned to the erroneous variable. The appearance of close values is an indication of BAD SET errors.

- NEG REGS: The presence of large negative values, which support BAD ADJUST and BAD LOOP errors.

- BORDER PROXIMITY: A measure of how close to the end of a valid memory block is being addressed by actions along a path, e.g., a variable initialized prior to a loop entry to point near the end of a block indicates a faulty initialization, particularly if the loop construction itself shows no signs of errors.

A sample inference with a PASTHELD belief network is displayed in Figure 5. In this case, DAACS generates 10 feasible paths to the error. Thus, before we collect path-specific evidence, the prior probability of each path for harboring the error is assumed to be 0.1. Figure 5 shows the result of an analysis of path 6, a path with significant probability. The path analysis revealed that:

1. SYNTACTIC STRUCTURE = *Set Adjust in Loop Fail*

2. NEG REGS = *False*, and

3. CLOSE REGS = *False*.

The probability that the error lies on path 6, given this information is 0.70, with a breakdown of 0.42 for a BAD SET and 0.28 for a BAD LOOP. Other problems are ruled out on the path.

While the system uses a rather simplistic approach to probabilistic reasoning, it is an extension to the current practice of logical and rule based analysis. The analysis provided by the system, in conjunction with a built in graphical program visualization tool, helps guide the programmer in the search for the error and trains novice to intermediate-skilled programmers in debugging techniques. Expert debuggers estimate that the tool will reduce debugging time by 25-50%. Moreover, preliminary metrics indicate that the systems can successfully diagnose 10% more dumps than by using logical analysis alone. The explanation of findings by path has shown to be more natural, and hence more understandable, presentation of the results than the previous version of the system.

## 6  UNCERTAINTY AND PATH IDENTIFICATION

We are working to embed uncertain reasoning machinery in the path-identification phase of DAACS. We believe that interleaving a probabilistic analysis into the path identification procedures so as to consider the likelihood that a path–which cannot be refuted with a logical analysis–is irrelevant to an error will be useful. Determining the likelihood of execution paths can be difficult because we rely on incomplete state of information about processes leading to an error. One source of incompleteness is based on the common problem that temporal sequences of computational actions can destroy valuable evidence about execution paths. If an address cannot be derived with certainty, we cannot derive an execution path with certainty. We are examining the use of information about the structure of a program, including such information as classes of interaction at branching points, to reason about likelihoods that different candidate paths were the actual route taken on the way to an identified error. We hope to automate the computation of probabilities $p(path\ executed\ |\ structure\ of\ program)$. This would allow us to relax our assumption that all valid paths have equal prior probabilities of being the fateful erroneous path that had been taken by the program. Such analyses hold promise for allowing us to make use of statistical information gathered by autonomous operating-system monitoring agents, which might be designed to focus their attention on relevant components of a system's operation.

Uncertain reasoning methods in the identification of paths may be especially important in scaling-up program analyses beyond a focus on program segments. Several probabilistic and decision-theoretic tools may be useful for grappling with the complexity of analyses that consider interactions among multiple program

```
                    o
                    o
                    o
• PATH 4: p(ERROR)=.01

• PATH 5: p(ERROR)=.05
________________________________________

• PATH 6

              PATH FINDINGS

> SYNTACTIC STRUCTURE: Set Adjust in Loop Fail
> NEG REGS: False
> CLOSE REGS: False

           PATH FAILURE PROBABILITY

         p(ERROR ON PATH 6): 0.70

              ERROR BREAKDOWN
              > BAD SET:  0.42
              > BAD LOOP: 0.28
                    o
                    o
________________________________________

• PATH 7: p(ERROR)=.08

• PATH 8: p(ERROR)=.02

                    o
                    o
                    o
```

Figure 5: Sample inference of a PASTHELD Error. Path 6 is expanded to display a breakdown of the likelihoods of different classes of problem on that path.

segments. For one, we are interested in exploring the value of integrating a higher-level uncertainty analysis, with the path-relevance and path-error analyses. As we mentioned in the introduction, early on, we developed a belief network for reasoning about different scenarios that could explain a problem, by looking at information about the application and salient variables. We speculate that we may be able to take advantage of statistical and expert knowledge about program failure at a higher level of detail to use in focusing the uncertainty analyses at the path-identification and path-relevance levels. For example, a global uncertainty analysis might be useful for computing the prior probabilities of alternative faults that will be used for each path analysis. Also, experts may be comfortable building and assessing path-error-analysis belief networks conditioned on different high-level explanations for a problem. Additionally, we are interested in the use of decision-theoretic methods to focus the attention of path-identification analyses, to identify cost-effective evidence- gathering strategies, and to prioritize debugging tasks for a time-pressured software engineer (Horvitz, 1988; Horvitz and Rutledge, 1991).

## 7  SUMMARY

We described our approach to software diagnosis based on an integration of logical program understanding for generating a set of feasible paths, and uncertain reasoning for determining the likelihood that problems are on specific paths. We plan to move beyond the use of purely logical methods for identifying paths, and thus, to relax the assumption about the equivalent relevance of paths and partial paths to software problems. We also intend to initiate studies to evaluate the fidelity of the methodology by investigating the calibration of the likelihoods computed by the system, and to investigate the time savings gained by engineers who take advantage of the prioritization of software paths recommended by DAACS-II.

## References

Arbon, R., Atkinson, L., Chen, J., and Guida, C. (1992). Tpf dump analyzer: A system to provide expert assistance to analysts in solving run-time program exceptions by deriving program intention from a tpf assembly language program. In *Proceedings of Fourth Conference on Innovative Application of Artificial Intelligence*, pages 71–88, Menlo Park, California. American Association for Artificial Intelligence.

Burnell, L. and Talbot, S. (1993). Incorporating probabilistic reasoning in a reactive program debugging system. In *Proceedings of the Ninth IEEE Conference on Artificial Intelligence for Applications*, pages 321–327. IEEE.

Hartman, J. (1992). Technical introduction on automated program understanding. In *AAAI Workshop on Artificial Intelligence and Automated Program Understanding, San Jose, California*.

Horvitz, E. (1988). Reasoning under varying and uncertain resource constraints. In *Proceedings AAAI-88 Seventh National Conference on Artificial Intelligence, Minneapolis, MN*, pages 111–116. Morgan Kaufmann, San Mateo, CA.

Horvitz, E. and Rutledge, G. (1991). Time-dependent utility and action under uncertainty. In *Proceedings of Seventh Conference on Uncertainty in Artificial Intelligence, Los Angeles, CA*. Morgan Kaufman, San Mateo, CA.

Kozaczynski, W., Letovsky, S., and Ning, J. (1991). A knowledge-based approach to software system understanding. In *Proceedings of the Sixth Annual Knowledge-Based Software Engineering Conference*, pages 162–170.

Lenz, N. and Saelens, S. (1991). A knowledge-based system for MVS dump analysis. *IBM Systems Journal*, 30(3):336–350.

Selfridge, P. (1991). Knowledge representation support for a software information system. In *Proceedings of the Seventh IEEE Conference on Artificial Intelligence for Applications*, pages 134–140. IEEE.